\documentclass[conference]{IEEEtran}
\IEEEoverridecommandlockouts
\def\etal{\textit{et al.}}
\usepackage{amsmath,amssymb,amsfonts}
\usepackage{algorithmic}
\usepackage{graphicx}
\usepackage{textcomp}
\usepackage{xcolor}
\usepackage{hyperref}

\usepackage{caption}
\usepackage{multirow}
\usepackage{amssymb}
\usepackage{pifont}
\newcommand{\xmark}{\ding{55}}%

\def\BibTeX{{\rm B\kern-.05em{\sc i\kern-.025em b}\kern-.08em
    T\kern-.1667em\lower.7ex\hbox{E}\kern-.125emX}}
\begin{document}

\title{Graph Contrastive Learning for Connectome Classification
\thanks{This work was partially funded by CSIC (I+D project 22520220100076UD) and ANII (FCE-1-2023-1-176172). Both Schmidt and Silva contributed equally to this work and share first authorship. }

}

\author{\IEEEauthorblockN{
Mart\'in Schmidt\IEEEauthorrefmark{2}\IEEEauthorrefmark{1},
Sara Silva\IEEEauthorrefmark{1},
Federico Larroca\IEEEauthorrefmark{1},
Gonzalo Mateos\IEEEauthorrefmark{2}
and
Pablo Mus\'e\IEEEauthorrefmark{1}}
\IEEEauthorblockA{\IEEEauthorrefmark{1}Facultad de Ingenier\'ia, Universidad de la Rep\'ublica, Uruguay
}
\IEEEauthorblockA{\IEEEauthorrefmark{2}Dept. of Electrical and Computer Engineering, University of Rochester, Rochester, NY, USA.\\
Emails: \{ssilva,flarroca,pmuse\}@fing.edu.uy and \{mschmi21,gmateosb\}@ur.rochester.edu}
}


%
\maketitle

\begin{abstract}


With recent advancements in non-invasive techniques for measuring brain activity, such as magnetic resonance imaging (MRI), the study of structural and functional brain networks through graph signal processing (GSP) has gained notable prominence. GSP stands as a key tool in unraveling the interplay between the brain’s function and structure, enabling the analysis of graphs defined by the connections between regions of interest—referred to as connectomes in this context. Our work represents a further step in this direction by exploring supervised contrastive learning methods within the realm of graph representation learning. The main objective of this approach is to generate subject-level (i.e., graph-level) vector representations that bring together subjects sharing the same label while separating those with different labels. These connectome embeddings are derived from a graph neural network Encoder-Decoder architecture, which jointly considers structural and functional connectivity. By leveraging data augmentation techniques, the proposed framework achieves state-of-the-art performance in a gender classification task using Human Connectome Project data. More broadly, our connectome-centric methodological advances support the promising prospect of using GSP to discover more about brain function, with potential impact to understanding heterogeneity in the neurodegeneration for precision medicine and diagnosis.

\end{abstract}

\section{Introduction}

The study of human brain connectivity, both structural and functional, is essential to decipher the brain's pivotal role in regulating behavior and cognitive processes. Neuroimaging advances, such as functional magnetic resonance imaging (fMRI) and positron emission tomography (PET), enable real-time visualization of brain activity, offering valuable insights into functional connectivity (FC)~\cite{van2010exploring}. Furthermore, structural mapping through techniques such as diffusion magnetic resonance imaging (dMRI) facilitates the construction of macroscopic maps of the physical connections of the brain~\cite{fornito2016fundamentals}. These tools have been instrumental in the diagnosis and treatment of neurological and psychiatric conditions, including epilepsy, autism, and schizophrenia \cite{esquizofrenia, epilepsia, autismo}.

This work focuses on Contrastive Learning (CL) techniques applied to Graph Neural Networks (GNNs), which excel in processing data structured as graphs; see e.g.,~\cite{gama2020graphs}. CL is a widely used machine learning technique that learns robust representations by enforcing similarity between related data points while separating unrelated ones. CL approaches, e.g., the SimCLR framework~\cite{SimCLR}, have achieved significant success in fields such as computer vision. However, only a few CL studies including~\cite{GCLAugmentations, empiricalGCL, fairCL} have focused on graph-structured data, leaving this area underexplored in the literature. In particular, we employ these techniques for subject-level (gender) classification using FC and structural connectomes (SC) extracted from fMRI and dMRI data, respectively. Connectomes are naturally modeled as graphs, where each node corresponds to a region of interest (ROI) and each edge represents the neural connections between ROIs, thereby capturing their intricate and complex relationships~\cite{sporns2010networks}. Beyond the specific gender classification task, we aim to contribute in the pathway to advance our understanding of brain networks and develop tools with broader applications in neuroscience. \vspace{2pt}

\noindent \textbf{Methodological innovation.} A GNN may be understood as a concatenation of several layers, each consisting of a graph convolution followed by a point-wise non-linearity~\cite{gama2020graphs}. Graph convolutions are fundamental tools of Graph Signal Processing (GSP)~\cite{ortega2018graph}, a framework that has recently gained prominence for the analysis of brain networks and associated signals~\cite{huang2016graph,huang2018graph}. 
The rationale behind applying the CL technique to our problem is inspired by Li \etal ~\cite{li2022learning}, who initially implemented the GNN architecture used in this study for connectome classification and later proposed a framework incorporating concepts akin to CL~\cite{li2022similarity}. Building on this foundation, our aim is to structure the latent space such that learned representations capture meaningful relationships within the data, ultimately enhancing classification performance. Different from~\cite{li2022learning,li2022similarity}, we propose a novel two-stage GNN training scheme. In the first stage, the pre-training phase, the model learns
to minimize a supervised similarity-preserving loss, whereas in the fine-tuning phase, a classifier is incorporated into the pipeline for
an additional training step. Furthermore, we integrate data augmentation techniques to generate diverse views of the same subject, ensuring that the model achieves a closer alignment between these augmented versions during the pre-training phase.\vspace{2pt} 

\noindent \textbf{Evaluation protocol.} To evaluate the effectiveness of this approach, we compare against a more conventional graph classification framework, where a GNN is trained directly for the  task without the additional CL step. This comparison provides insights into the potential advantages of CL in graph-based modeling and its impact on connectome classification performance. Through this study, our work contributes to the growing exploration of advanced representation learning techniques in network neuroscience~\cite{ktena2018metric,sihag2023explainable,neurograph}.

Moreover, to assess the robustness of the CL framework and to evaluate the impact of data quantity on the performance of each proposed model, we conducted experiments using reduced versions of the training dataset. This investigation is particularly pertinent in the medical domain, where acquiring large-scale datasets is often challenging due to patient privacy concerns, high data collection costs, and the inherent rarity of certain neurological conditions. Consequently, researchers often face the need to develop effective models under data scarcity. \vspace{2pt}

\noindent \textbf{Data.} The dataset used in the present work is part of the Human Connectome Project (HCP)~\cite{HCP, HCP_DC}, a large-scale initiative aimed at mapping the human connectome with advanced MRI techniques. Specifically, we utilized the HCP Young Adults 1200 release, which includes data from 1200 healthy adults.\footnote{Visit the project website \url{https://www.humanconnectome.org/study/hcp-young-adult} and the repository
\url{https://www.humanconnectome.org/study/hcp-young-adult/document/1200-subjects-data-release} for details.} This release provides comprehensive MRI data, along with demographic and behavioral information. For this study, we derived FC matrices from functional MRI (fMRI) timecourses and SC from diffusion MRI (dMRI) data using the Desikan-Killiany atlas~\cite{DESIKAN2006968}. After preprocessing, the dataset includes 1048 subjects with complete data, as not all participants underwent both imaging modalities. This rich dataset enables a robust exploration of brain connectivity and its relationship to individual characteristics.

All in all, our main contributions are the following:\vspace{2pt}  

    \noindent $\bullet$ \textbf{Evaluation of contrastive learning methods}: We evaluate the impact of applying CL techniques to improve the quality of learned representations for connectome classification, including experiments with reduced training sets to emulate the common scenario of limited data in medical applications. 
    
    \noindent $\bullet$  \textbf{Effect of data augmentation}: We analyze the role of data augmentation strategies within the CL framework and their influence on model performance.  
    
    \noindent $\bullet$  \textbf{Integration with Encoder-Decoder architectures}: We explore the combination of CL with the Encoder-Decoder approach for SC and FC in~\cite{li2022learning}, building on the architecture therein to enhance representation learning.  
    
    \noindent $\bullet$ \textbf{State-of-the-art classification}: We demonstrate competitive or superior classification performance relative to state-of-the-art methods, benchmarking against a key study~\cite{neurograph}.\vspace{2pt}   

\noindent \emph{Notation:} In what follows, matrices are represented by bold uppercase letters (\(\mathbf{X}\)), column vectors by bold lowercase letters (\(\mathbf{x}\)) and plain letters ($x$) denote scalars or indices. The Frobenius norm of a matrix \(\mathbf{M} \in \mathbb{R}^{N \times N}\) is defined as $\|\mathbf{M}\|_F^2:= \sum_{i=1}^N\sum_{j=1}^N  M_{ij}^2.$ The remaining notation is explicitly defined when introduced.

\section{Methods}

This section presents an overview of the proposed methodology. We begin by describing the Encoder-Decoder baseline model in Section \ref{sec:ed}, and then we introduce two useful data augmentation techniques (Section \ref{sec:da}). Finally, Section \ref{sec:cl} outlines how we integrate these techniques into a supervised CL framework rooted in the Encoder-Decoder architecture.

\subsection{Encoder-decoder architecture}
\label{sec:ed}

\begin{figure*}
\centering
\includegraphics[width=\textwidth]{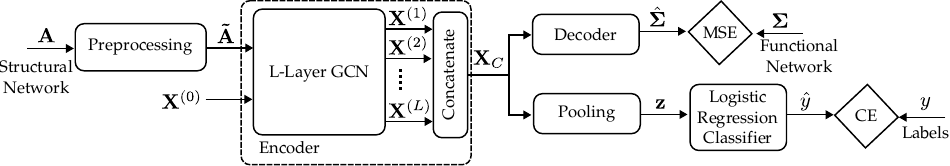}
\caption{
The Encoder-Decoder architecture used in this work processes structural connectivity matrix \(\mathbf{A}\) as input, along with nodal attributes \(\mathbf{X}^{(0)} = \mathbf{I}_{N}\). The GCN encoder performs graph convolutions and associated information propagation through $L$ layers, producing low-dimensional node embeddings \(\mathbf{X}^{(l)}\) at layer $l$. Intermediate representations from all layers are concatenated into one final embedding \(\mathbf{X}_C\). The decoder uses an outer-product operation to predict the reconstructed functional connectivity matrix \(\hat{\mathbf{\Sigma}}:=\text{ReLU}(\mathbf{X}_C\mathbf{X}_C^\top)\), effectively modeling the SC-FC mapping. Graph-level representations \(\mathbf{z}\) are obtained via a pooling operation applied to \(\mathbf{X}_C\), summarizing the node embeddings into a single vector, from which a logistic regression classifier predicts the binary label \(\hat{y}\).\vspace{2pt} 
}
\label{fig:encoder-decoder}
\end{figure*}

The multi-task Encoder-Decoder architecture used in this work is illustrated in \autoref{fig:encoder-decoder}. This model consists of two main components: the top branch performs a regression task to reconstruct the FC matrix from the SC matrix, while the bottom branch predicts the gender of the subject using a graph classification module. 
These components are tightly integrated to facilitate learning of latent representations with inductive biases beneficial to both tasks.

The following are key components of the architecture. 

    \noindent \textbf{Inputs.}  
    The model takes as input the SC matrix \(\mathbf{A} \in \mathbb{R}^{N \times N}\), the FC matrix \(\mathbf{\Sigma} \in [0, 1]^{N \times N}\), and an initial nodal feature matrix \(\mathbf{X}^{(0)} = \mathbf{I}_{N}\) (the identity matrix; i.e., a one-hot encoding vector per ROI), where $N$ is the number of ROIs (with $N=87$, as per the Desikan-Killiany atlas). \vspace{2pt}

    \noindent \textbf{GCN encoder.}  
    The encoder uses a multi-layer graph convolutional network (GCN) to generate low-dimensional node embeddings. At each layer \(l\), the GCN applies a graph convolution followed by a $\textrm{ReLU}(\cdot)=\max(\cdot,0)$ non-linearity:  
    \[
    \mathbf{X}^{(l+1)} = \text{ReLU}(\tilde{\mathbf{A}}\mathbf{X}^{(l)}\mathbf{\Theta}^{(l)}),
    \]
    where \(\tilde{\mathbf{A}}\) is the normalized SC adjacency matrix, as defined in~\cite{primer_GCN}, \(\mathbf{X}^{(l)} \in \mathbb{R}^{N \times d_{l-1}}\) is the feature matrix at layer \(l\), \(\mathbf{\Theta}^{(l)} \in \mathbb{R}^{d_{l-1} \times d_l}\) are the trainable weights and \( d_l \) denotes the dimensionality of the node embeddings at layer \( l \). The decoder output is a concatenated embedding \(\mathbf{X}_C \in \mathbb{R}^{N \times \sum_l d_l}\), combining the representations from all layers $l=1,\ldots,L$.\vspace{2pt}

    \noindent \textbf{Decoder.}  
    The decoder reconstructs the FC matrix from \(\mathbf{X}_C\) as \(\hat{\mathbf{\Sigma}} :   = \text{ReLU}(\mathbf{X}_C \mathbf{X}_C^\top)\). This step ensures that \(\hat{\mathbf{\Sigma}} \in \mathbb{R}^{N \times N}\) has positive entries, aligning with the FC reconstruction objective. The reconstruction error is measured using the Mean Squared Error (MSE) between the entries of \(\hat{\mathbf{\Sigma}}\) and the ground truth FC matrix \(\mathbf{\Sigma}\).\vspace{2pt}

    \noindent \textbf{Pooling and classification.}  
    For the classification branch, a pooling layer aggregates \(\mathbf{X}_C\) across nodes by averaging, resulting in a graph-level embedding \(\mathbf{z}\in \mathbb{R}^{\sum_l d_l}\). This embedding is then fed to a logistic regression classifier to predict the subject's gender, using a binary cross-entropy (CE) loss.\vspace{2pt} 

    \noindent \textbf{Loss function.}  
    The model is trained using a joint reconstruction and classification loss:  
\[
\mathcal{L}
\;=\;
\underbrace{\mathcal{L}_{\mathrm{MSE}}\bigl(\hat{\mathbf{\Sigma}}, \mathbf{\Sigma}\bigr)}_{\text{reconstruction loss}}
\;+\;
\lambda
\,\times\,
\underbrace{\mathcal{L}_{\mathrm{CE}}\bigl(\hat{y}, y\bigr)}_{\text{classification loss}},
\]
where \(\mathbf{\Sigma}\) is the ground truth FC matrix, \(y\) is the ground truth gender label, and \(\hat{\mathbf{\Sigma}}\) and \(\hat{y}\) are their respective predictions. The hyperparameter \(\lambda\) balances the importance of the reconstruction and classification objectives.
    The reconstruction MSE is computed as:
    \begin{equation}
    \begin{aligned}
    \mathcal{L}_{\mathrm{MSE}}\bigl(\hat{\mathbf{\Sigma}}, \mathbf{\Sigma}\bigr)
    \;=\;
    \frac{1}{N^2} \,\bigl\|\hat{\mathbf{\Sigma}} - \mathbf{\Sigma}\bigr\|_{F}^{2}.
    \end{aligned}
    \label{eq:l_mse}
    \end{equation}
    On the other hand, minimizing the classification loss aims at predicting the binary label \(y\) given \(\hat{y}\). We adopt the binary CE loss, which for a single sample amounts to:
    \begin{equation}
    \begin{aligned}
    \mathcal{L}_{\mathrm{CE}}\bigl(\hat{y}, y\bigr)
    \;=\;
    -\,\Bigl[y\,\log\!\bigl(\hat{y}\bigr)
    \;+\;(1-y)\,\log\!\bigl(1-\hat{y}\bigr)\Bigr].
    \end{aligned}
    \label{eq:l_ce}
    \end{equation}
When training on a batch, both loss functions are averaged across all samples.


This dual-task pipeline enables the model to learn robust graph representations that are effective for reconstructing FC matrices and predicting gender (or other subject-level binary attributes). Even if the end goal is purely discriminative, the regression branch implicitly infuses the embeddings $\mathbf{X}_C$ with valuable information about the SC-FC coupling~\cite{li2022learning}.

\subsection{Data augmentation for graphs} \label{sec:da}

Data augmentation is widely used in deep learning to address the challenge of limited data availability and improve generalization. Two graph augmentation strategies were considered in this work, which we now briefly discuss, noting that other strategies are possible~\cite{DAGraphs, GCLAugmentations}.

In \emph{attribute masking} certain node attributes (i.e., the columns of $\mathbf{X}^{(0)}$) are masked by setting them to zero. Recall that the one-hot encoding we are using as input signal acts as an identifier for the ROI.
    By setting the corresponding signal vector to zero, 
    we are challenging the model to infer informative representations 
    ignoring some nodes' identities. 

\emph{Edge dropping} may be regarded as the dual of attribute masking, where now edges $A_{ij}$ of the SC graph are randomly dropped based on a probability \(p\). This forces the model to remain robust to missing links and generate consistent representations despite an incomplete graph structure.

In the CL framework, augmented versions of the data are incorporated to encourage the model to produce meaningful representations, even when the input data is perturbed. These augmentations aim to preserve the overall structure of the data while introducing variations that can aid generalization.

\subsection{Supervised contrastive learning}\label{sec:cl}

\begin{figure}
    \centering
    \includegraphics[width=1\linewidth]{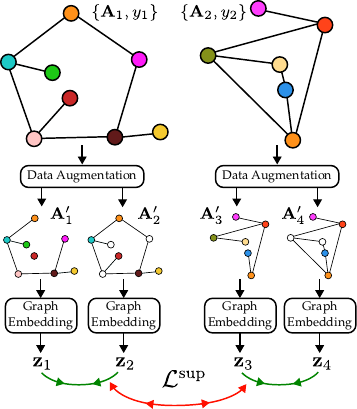}
    \caption{A schematic view of the constrastive learning approach. For each input datum $\{\mathbf{A}_k,y_k\}$ we generate two augmented versions $\{\mathbf{A}'_{2k},y_k\}$ and $\{\mathbf{A}'_{2k+1},y_k\}$ using edge dropping and attribute masking. The corresponding graph embeddings $\mathbf{z}_i$ are produced as in Fig.\ \ref{fig:encoder-decoder}. Whereas unsupervised CL only pulls together the representations stemming from the same original graph (represented by green arrows), supervised CL optimizes the contrastive loss $\mathcal{L}^{\mathrm{sup}}$ in \eqref{eq:l_sup_loss}, which also takes into account the class labels of the subjects. This approach attracts augmented versions originating from subjects with the same label and repels those associated with different ones (represented by green but also red arrows).}

    \label{fig:contrastive_learning}
\end{figure}

As illustrated in \autoref{fig:contrastive_learning}, the goal of CL is to train models to learn vector representations that are interpretable or useful for downstream tasks. The core principle is to ensure that similar data points are mapped to similar representations. Supervised CL, introduced in~\cite{supervisedCL}, extends this approach to a fully supervised framework. In this setting, not only are augmented versions of the same data point pulled closer together and distinct data points pushed apart, but this process is also guided by class labels: data points of the same class are attracted to each other, while those from different classes are repelled.

Given a batch of \(B\) data points and their corresponding labels \(\{\mathbf{A}_k, y_k\}_{k=1}^{B}\), the training batch consists of \(2B\) pairs \(\{\mathbf{A}'_k, y'_k\}_{k=1}^{2B}\). Here, \(\mathbf{A}'_{2k-1}\) and \(\mathbf{A}'_{2k}\) are two augmented versions of the original SC sample \(\mathbf{A}_k\), generated with attribute masking and edge dropping respectively (see Section \ref{sec:da}), and \(y'_{2k-1} = y'_{2k} = y_k\).
We propose a training process consisting of two steps: pre-training and fine-tuning.\vspace{2pt}

\noindent \textbf{Pre-training.} During the pre-training phase, the architecture is the one presented in Section \ref{sec:ed}, albeit without the logistic regressor (i.e., no classification is performed). In this step, the objective is to minimize the contrastive loss:
\small\begin{gather}
    \mathcal{L}^{\mathrm{sup}}\bigl(\{\mathbf{z}_k, y_k\}_{k=1}^{2B}\bigr) = \sum_{i \in I} \frac{-1}{|P(i)|} \sum_{p \in P(i)} \log \frac{\exp\bigl(\frac{\mathbf{z}_i^\top \mathbf{z}_p}{\tau}\bigr)}{\sum_{q \in Q(i)} \exp\bigl(\frac{\mathbf{z}_i^\top \mathbf{z}_q}{\tau}\bigr)},
   \label{eq:l_sup_loss}
\end{gather}\normalsize
%
where \(\mathbf{z}_i\) is the vector representation of matrix \(\mathbf{A}'_i\) (cf.\ after pooling in the lower branch in \autoref{fig:encoder-decoder}),  \(I = \{1, \ldots, 2B\}\) represents the set of indices in the augmented batch, \(Q(i) = I \setminus \{i\}\) is the set of indices excluding \(i\), \(P(i) = \{p \in Q(i) : y'_p = y'_i\}\) is the set of positive indices distinct from \(i\) that share the same label, and \(\tau \in (0,1]\) is a temperature hyperparameter for the model.

When the decoder branch is integrated to this framework, the model is trained using a combination of $\mathcal{L}^{\text{sup}}$ in \eqref{eq:l_sup_loss} and the reconstruction loss $\mathcal{L}_{\mathrm{MSE}}$ in~\eqref{eq:l_mse}. In this case, the reconstructed FC matrix $\hat{\mathbf{\Sigma}}_k$ is generated from the non-augmented SC matrix $\mathbf{A}_k$. The overall loss function in pre-training is therefore defined as follows:
\[
\mathcal{L}
\;=\;
\frac{1}{B} \sum_{k=1}^{B} \mathcal{L}_{\mathrm{MSE}}\bigl(\hat{\mathbf{\Sigma}}_k, \mathbf{\Sigma}_k\bigr)
\;+\;
\lambda
\,\times\,
\mathcal{L}^{\mathrm{sup}}\bigl(\{\mathbf{z}_k, y_k\}_{k=1}^{2B}\bigr).
\]
Note that in this step, we only adapt the parameters of the GCN encoder since the pooling that produces $\mathbf{z}$ from $\mathbf{X}_C$ and the decoder branch do not have trainable parameters. \vspace{2pt}

\noindent \textbf{Fine-tuning.} After the pre-training step, we turn to the fine-tuning phase, where the logistic regression classifier is incorporated into the pipeline and trained along with the encoder (cf.\ the lower branch in \autoref{fig:encoder-decoder}). In this step, the model is trained during \(K\) epochs using the classification loss function $\mathcal{L}_\mathrm{CE}$ in \eqref{eq:l_ce}, omitting the reconstruction loss $\mathcal{L}_{\mathrm{MSE}}$. During the first \(M < K\) epochs, only the classifier is trained (the encoder remains frozen). In the subsequent \(K - M\) epochs, the entire network (i.e., the GCN and the classifier) is trained  using a lower learning rate. This step refines the embeddings generated by the encoder so that they are more suitable for the specific classification task.

\section{Experiments}

\subsection{Implementation details}

For the gender classification task, we tested two architectures: Encoder-only and Encoder-Decoder. The latter is the one depicted in \autoref{fig:encoder-decoder}, whereas the former does not include the decoder branch (thus ignoring the FC). These architectures are evaluated on the classification task, both in a baseline setting (i.e., without a pre-training CL step) and in a supervised CL one. In addition, each framework was trained both with and without data augmentation, resulting in four final configurations:
\begin{itemize}
    \item Baseline setup without any modifications, model and loss as described in Section \ref{sec:ed}.
    \item Baseline setup with data augmentation, in which the model described in Section \ref{sec:ed} is trained on augmented versions of the SC matrix; see Section \ref{sec:da}.
    \item Contrastive learning with data augmentation, where the model is as described in Section \ref{sec:cl}. 
    \item Contrastive learning without data augmentation, where the model is as described in Section \ref{sec:cl}, but no data augmentation is applied.  
\end{itemize}

As described in Section \ref{sec:da}, the data augmentation techniques used were attribute masking and edge dropping, where between 0 and 87 nodes were masked and edges were removed with a probability of $p= 0.2$. In the non-CL framework, we randomly sampled one type of data augmentation method in each batch.

We performed a hyperparameter search, resulting in $L=3$ layers and encoder layer widths of $[32, 16, 8]$ for both architectures and frameworks. The learning rate used in the baseline classification was set to $0.001$, the batch size $B=64$, and the trade-off parameter $\lambda=0.4$. For the CL framework we set $\lambda= 0.25$, while the rest of the setup is as follows:
\begin{enumerate}
    \item During pre-training, a batch size of $B=128$, 3000 epochs, a learning rate of $1 \times 10^{-3}$, and $\tau=1$. 
    \item During fine-tuning, the encoder was frozen for the first set of $M=100$ epochs, and the learning rate was set to $1 \times 10^{-3}$. For the additional $K-M=100$ epochs, the complete model was further trained with a learning rate of $1 \times 10^{-4}$. This step is trained with a batch size of $B=16$.
\end{enumerate}

In all cases, the data was split into 80\% for training, 10\% for validation, and 10\% for testing. The reported results are based on the test set metrics. 

The complete code is available at \url{https://github.com/sara-silvaad/Connectome_GCL}.

\subsection{Results}

Results are displayed in \autoref{tab:encoder}. The first key observation is that the decoder branch improves performance. Specifically, using an Encoder-Decoder scheme outperforms the encoder-only approach in most cases. This improvement could be justified by the incorporation of the FC matrix, which provides richer information, and by the fact that the decoder acts as a form of ``regularizer” for the latent representation used for the classification task.

\begin{table}
\caption{Performance of the Encoder and Encoder-Decoder architectures under different frameworks: Baseline classification and contrastive learning (CL), both trained with and without data augmentation. The Decoder branch is beneficial in most cases. Data augmentation proves crucial within the CL framework, achieving the best performance overall. }
\label{tab:encoder}
\begin{center}
\renewcommand{\arraystretch}{1.5} 
\begin{tabular}{|l|c|c|c|}
\hline
\textbf{\multirow{2}{*}{Framework}} & \textbf{\multirow{2}{*}{Data Augmentation}} & \multicolumn{2}{c|}{\textbf{Architecture}} \\ \cline{3-4} 
                           &                                    & \textbf{Encoder} & \textbf{Encoder-Decoder} \\ \hline
Baseline class.                &  \xmark
                                  & 0.88                      & 0.90 \\ \hline
Baseline class.                &  \checkmark \par
                                  & 0.88                      & 0.90 \\ \hline
CL                       &  \xmark
                                  & 0.86                      & 0.86      \\ \hline
CL                        &  \checkmark \par
                                  & 0.93                      & 0.94 \\ \hline
\end{tabular}
\end{center}
\end{table}

Moreover, data augmentation plays a crucial role in the CL framework, and when applied within this context, it yields a notable impact compared to its marginal effect in the baseline setup. As suggested in~\cite{supervisedCL}, the supervised CL framework outperforms the baseline, and the inclusion of data augmentation---as highlighted in~\cite{GCLAugmentations}---strengthens the method’s performance. Although it is not entirely clear why the CL approach achieves such a pronounced benefit over the baseline, one possible explanation resides in the closer alignment between the augmented versions of the same subject, improving representation capacity and the robustness of the model. This observation might also elucidate the significance of data augmentation, since the method explictly relies on augmented versions of the same subject to strengthen the model. 

\begin{figure}
    \centering
        \centering
        \includegraphics[width=0.99\linewidth]{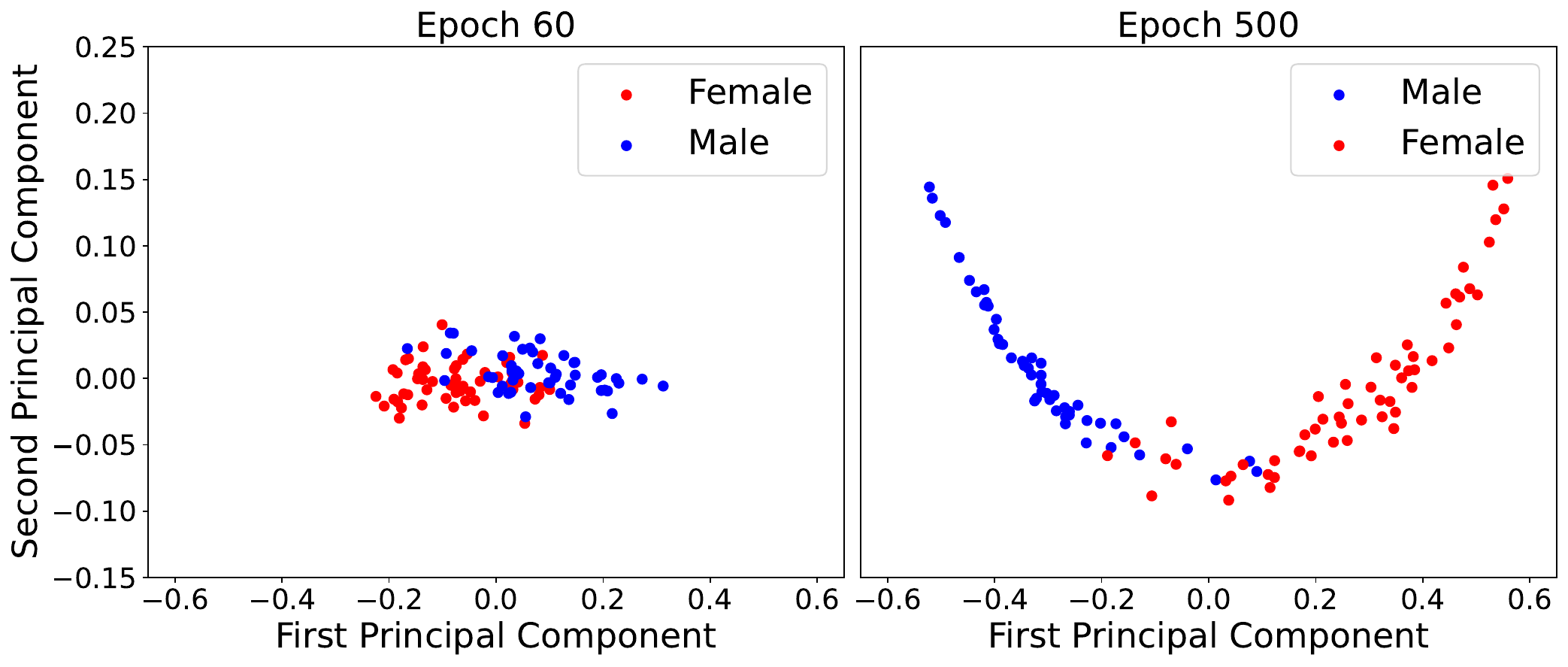}
    \caption{Evolution of the learned embeddings in the validation set for the contrastive learning setup and the Encoder-Decoder architecture, shown after the pre-training step. The first two PCA principal components are visualized. At epoch 500, in the first principal component, a clear separation between embeddings of different genders can be observed. During the fine-tuning step, these learned embeddings are utilized to classify with a logistic regression classifier.
    }
    \label{fig:pca}
\end{figure}

In any case, it is important to highlight that the combination of CL and data augmentation achieves the best performance: the best results were obtained using this scheme, surpassing~\cite{neurograph}, which, to the best of our knowledge, is the only work benchmarking the gender classification task using HCP subjects. Furthermore, \autoref{fig:pca} illustrates the evolution of the first two PCA dimensions of the learned representations throughout the epochs, for the most performant approach (Encoder-Decoder with CL and data augmentation). Notably, after the pre-training stage, a clear separation between embeddings of different classes can already be observed.\vspace{2pt}

\noindent \textbf{Robustness.} To analyze the impact of the sample size on the performance of the model, an evaluation was carried out using different proportions of the training set. The test set was kept invariant, while the training set was reduced to 10\%, 30\%, 50\% and 70\% of its original size to assess the robustness of the different frameworks. The batch size in the baseline setting was set to $B=8, 16, 32, 64$ respectively, while in CL: $B=16, 32, 64, 128$ for the pre-training step, and $B=4, 8, 16, 16$ for the fine-tuning. The rest of the hyperparameters kept the same values from the previous test case.

\begin{figure}
    \centering
    \includegraphics[width=1\linewidth]{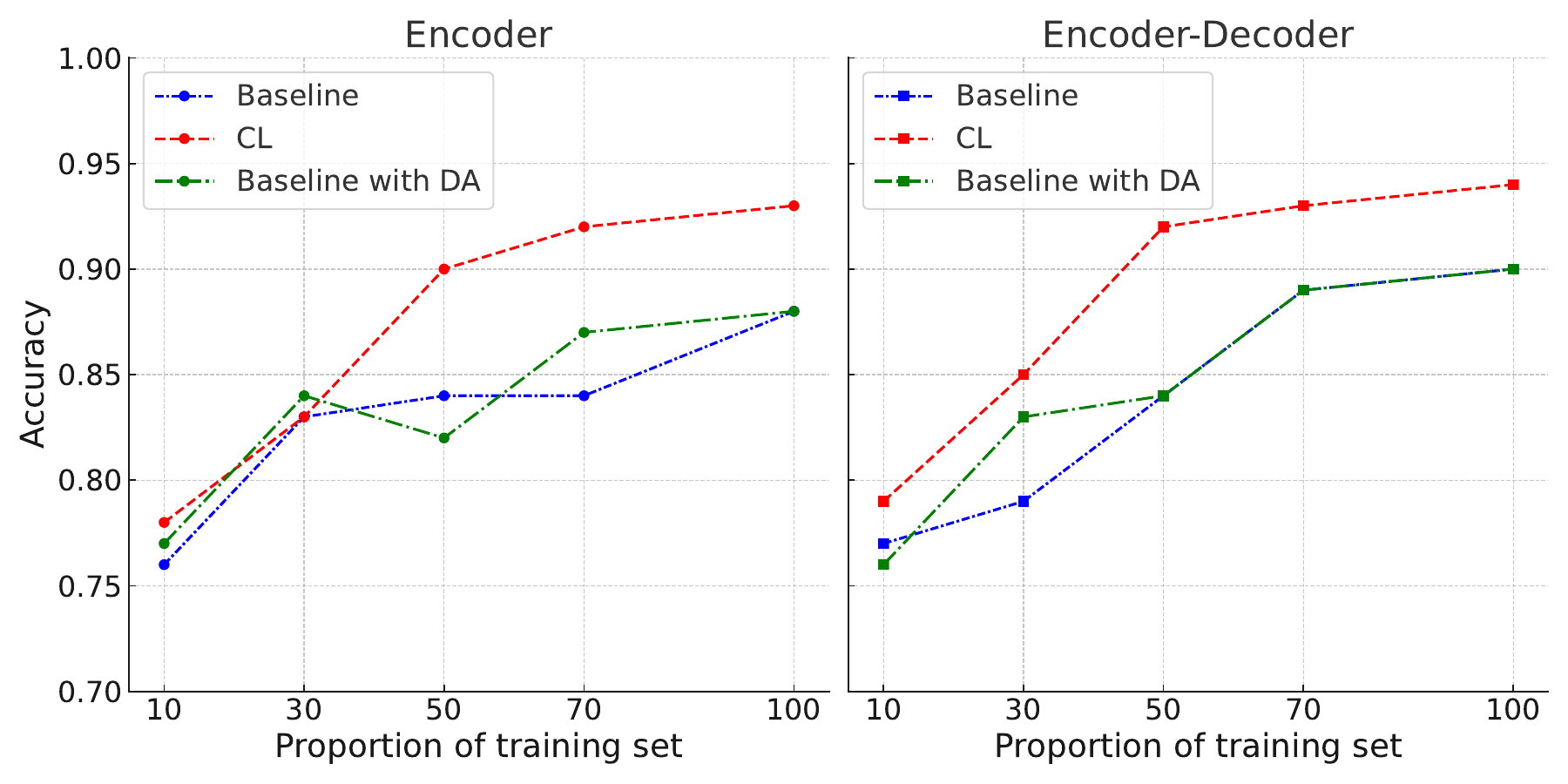}
    \caption{Comparison of contrastive learning (CL) and baseline classification (Baseline), for encoder and Encoder-Decoder architectures with or without data augmentation (DA), under varying training set proportions. The CL approach consistently outperforms the others and reaches near-optimal results with as little as half of the data, whereas the other frameworks require the full dataset to achieve near-peak performance.}
    \label{fig:prop}
\end{figure}

\autoref{fig:prop} shows the training results when using different proportions of the training set, comparing the performance between the different frameworks under varying data quantities. In all cases, across both architectures and all proportions of the training set, we find that the CL framework consistently outperforms all other frameworks, demonstrating its robustness even when the amount of available training data is limited. Interestingly, the CL framework starts to show strong performance with fewer subjects, attaining near-optimal accuracies with as few as half of the original data. On the other hand, competing alternatives require (almost) the full training set to reach their peak performance.

\section{Conclusion}

We have presented a CL methodology that not only obtains state-of-the-art results in subject classification, but also does so with markedly less data than competing alternatives.
Despite these promising preliminary findings, there remains a need to explore other data augmentation strategies, especially domain-specific ones (for instance, considering brain ROIs when dropping edges or masking attributes). It is important to acknowledge that current data augmentation strategies (i.e., attribute masking and edge dropping) can be detrimental, potentially removing critical information from the connectomes. Therefore, developing augmentation techniques that preserve vital brain connectivity patterns while introducing meaningful variability is a promising direction we will explore in the near future. 

Furthermore, as shown in \autoref{fig:pca}, the proposed methodology learns subject-level representations that are clearly separated after the pre-training step (when the subjects belong to different classes, naturally). 
This motivates conducting a more thorough analysis of the characteristics that distinguish subjects who are easier or harder to classify based on their distance in the embedding space. These characteristics may include demographic factors (e.g., age) or topological properties of the respective subjects' SC and FC matrices.

Overall, combining CL, data augmentation, and the Encoder-Decoder model is a promising approach to achieve robust and interpretable representations, especially for network neuroscience tasks.

\addtolength{\textheight}{-6cm}   










\bibliographystyle{IEEEtran}
\bibliography{referencias}

\end{document}